\documentclass[10pt, a4paper]{article}
\usepackage{lrec2022} 
\usepackage{multibib}
\newcites{languageresource}{Language Resources}
\usepackage{graphicx}
\usepackage{tabularx}
\usepackage{soul}
\usepackage{multirow}
\usepackage{titlesec}
\titleformat{\section}{\normalfont\large\bfseries\center}{\thesection.}{1em}{}
\titleformat{\subsection}{\normalfont\SmallTitleFont\bfseries\raggedright}{\thesubsection.}{1em}{}
\titleformat{\subsubsection}{\normalfont\normalsize\bfseries\raggedright}{\thesubsubsection.}{1em}{}
\renewcommand\thesection{\arabic{section}}
\renewcommand\thesubsection{\thesection.\arabic{subsection}}
\renewcommand\thesubsubsection{\thesubsection.\arabic{subsubsection}}

\usepackage{epstopdf}
\usepackage[utf8]{inputenc}

\usepackage{hyperref}
\usepackage{xstring}

\usepackage{color}
\usepackage{tipa}

\hypersetup{
           breaklinks=true,   
           colorlinks=true,   
           pdfusetitle=true,  
        }

\title{Huqariq: A Multilingual Speech Corpus of Native Languages of Peru for Speech Recognition}

\name{Rodolfo Zevallos, Luis Camacho, Nelsi Melgarejo} 

\address{Pompeu Fabra University, Pontifical Catholic University of Peru \\
         Barcelona Spain, Lima Peru \\
         rodolfojoel.zevallos@upf.edu, \{luis.camacho, nelsi.melgarejo\}@pucp.pe}

\abstract{
The Huqariq corpus is a multilingual collection of speech from native Peruvian languages. The transcribed corpus is intended for the research and development of speech technologies to preserve endangered languages in Peru. Huqariq is primarily designed for the development of automatic speech recognition, language identification and text-to-speech tools. In order to achieve corpus collection sustainably, we employ the crowdsourcing methodology. Huqariq includes four native languages of Peru, and it is expected that by the end of the year 2022, it can reach up to 20 native languages out of the 48 native languages in Peru. The corpus has 220 hours of transcribed audio recorded by more than 500 volunteers, making it the largest speech corpus for native languages in Peru. In order to verify the quality of the corpus, we present speech recognition experiments using 220 hours of fully transcribed audio.
 \\ \newline  }

\begin{document}

\maketitleabstract

\section{Introduction}

The Huqariq project responds to the endangerment currently faced by native languages in Latin America and the lack of language technologies faced by low-resource languages in Peru \cite{rogers2015endangered}. This situation is mainly due to the lack of speech corpora, which are the raw material for the creation of language tools, are scarce and the few that exist are privately licensed; for this reason, they should be in the public domain to contribute to the development and revitalization of languages. 

Around the world, there are some initiatives for the collection of corpora for low-resources languages and that are in the public domain that employ different methodologies and ways of collection. One of the most successful methodologies is crowdsourcing, i.e. native speakers volunteering to help in the construction of the resources. This methodology is supported by web tools or mobile applications, which can be massively used.

Our corpus collection tool is designed to expand organically to new native languages as community members record domain-specific base audios as prompts in the corpus collection. Unlike others (Common Voice \cite{ardila2019common}), this tool does not use text to be read, as many native speakers of indigenous languages are illiterate in their native language. Therefore, our tool replaces reading texts with listening to audios. This subtle but important change facilitates corpus collection.

\section{Prior work}

Although the majority of the speech corpora employed in the most widely used tools are private, there are some worldwide initiatives of speech corpora with open licenses and low-resource languages. In 2019 the VoxForge project \cite{VoxForge} collected a speech corpus for 17 languages; this project is community-driven in the same way Mozilla's Common Voice project \cite{ardila2019common} has collected 2500 hours of transcribed audio for 29 languages by crowdsourcing being one of the most community-supported projects.

On the other hand, the speech corpus projects of native languages of Latin America are almost null; in 2019, a speech corpus of 142 hours of fully transcribed Mapudungun was released \cite{duan2019resource}; in 2018, the siminchikkunarayku \cite{cardenas2018siminchik} project collected 99 hours of audio of southern Quechua. Unfortunately, both projects do not have open licenses.

\section{Native Languages}

Peru is a multicultural country, mainly due to the presence of native first nations, these make up a total of 10\% of the population. Because of this population there are still 48 native languages spoken, however, they are under the risk of extinction. These languages are facing some major issues like the lack of a unique grammar or writing system, lack of presence on the internet, lack of mass of expert linguists and lack of electronic resources. \cite{cardenas2018siminchik}
In this section we present some important linguistic characteristics relevant to NLP, especially in regards to dialectal and phonological variety which play an important role in speech based linguistic technology.

\subsection{Quechua}

Quechua (ISO 639-3 que) is a family of languages spoken in South America with about 10 million speakers, not only in the Andean regions but also in the valleys and plains connecting the Amazon jungle and the Pacific coast. Quechua languages are considered highly agglutinative with a subject-object-verb (SOV) sentence structure as well as mostly postpositional. Even though the classification of Quechua languages remains open to research \cite{heggarty,landerman}, recent work in language technology for Quechua \cite{rios2015basic,rios2014morphological} have adopted the categorization system described by Torero \cite{torero}. This categorization divides the Quechua languages into two main branches, QI (Glottolog quec1386) and QII (quec1388). Branch QI corresponds to the dialects spoken in central Peru, which are treated as one collective in this paper. QII is further divided in three branches, QIIA, QIIB and QIIC. QIIA groups the dialects spoken in Northern Peru, while QIIB the ones in Ecuador and Colombia. In this paper we work with QI (Central Quechua, Glottolog quec1386) and QIIC (Southern Quechua, Glottolog quec1389).

\subsubsection{Southern Quechua}

Southern Quechua (QIIC) has two main variants: Chanka Quechua (ISO 639-3 quy) and Collao Quechua, also known as Cusco Quechua (ISO 639-3 quz). In both dialects, only the vowels /a/, /i/ and /u/ are found as phonemic vowels. Referring to consonants, Chanka Quechua has a total of 15, most of them voiceless and as in Spanish, the phoneme /\textbottomtiebar{t\textesh}/ is written as \textit{ch}, /\textltailn/ as \textit{\~n}, and /\textturny/ as \textit{ll}. On the other hand, Collao Quechua also has a glottal and an aspirated version of each plosive consonant, giving it a total of 25 consonants. Both dialects have voiced consonants in their phonemic inventory due to the large number of borrowings from Spanish.

\subsubsection{Central Quechua}

Since there is greater dialectal variation among the variants of the QI branch compared to the variation between Quechua Chanka and Quechua Collao, we will go into a bit more detail in this section. Unlike Southern Quechua, Central Quechua (QI) has 3 short phonemic vowels /a/, /i/ and /u/, and 3 long phonemic vowels /aa/, /ii/ and /uu/. Central Quechua also has 3 nasal consonants /m/ /n/, /\textltailn/, 4 occlusive consonants /p/, /t/, /k/, /q/, 2 affricate consonants of variable value, 3 fricative consonants /s/, /\textesh/, /h/, 2 approximant consonants /j/, /w/ and 3 liquid consonants /$\lambda$/, /\textfishhookr/, /l/. The uvular /q/ is pronounced occlusive only in Callejón de Huaylas, being fricative in the other provinces: voiceless [\textchi] in Corongo for all positions, while in the Conchucos it is voiced [\textinvscr] in initial position and voiceless in coda. The alveolar nasal /n/ has three allophones, namely: the velar [\textltailn], in syllabic coda and when it precedes the velar [k], the uvular [\textscn] when it precedes [q], and the bilabial [m] before [p]. The vibrant [\textfishhookr] becomes a retroflex sibilant [\textcommatailz] at word onset. The voiced bilabial /b/, dental /d/ and velar /g/, as well as the voiceless bilabial fricatives /$\phi$/ and voiceless retroflex /\textcommatailz/, are used as distinct phonemes only in borrowings from Spanish. \cite{QC-escritura}.

\subsection{Aymara}

The Aymara language (ISO 639-3 aym) belongs to the Aru linguistic family, is spoken by the Aymara people and although it is in a vital state \cite{minedu2018}, it is considered an endangered language \cite{adelaar2014endangered}. Aymara is spoken in four countries: Argentina, Bolivia, Chile and Peru. In Peru, it is the second most spoken native language after Quechua, according to the 2017 census conducted by the National Institute of Statistics and Informatics \cite{INEI}.  

It is an agglutinative language.

Aymara has 3 short phonemic vowels /a/, /i/ and /u/, and 3 long \textit{ä} /aa/, \textit{ï} /ii/ and \textit{ü} /uu/. Also, it features 26 consonant phonemes, most of them aspirated occlusives \textit{ph} [$p^h$], \textit{th} [$t^h$] and \textit{kh} [$k^h$]. In addition, the aspirated postalveolar affricate is signaled by the triplet \textit{chh} [t\textesh$^h$] and an apostrophe is used to signal the occlusive and affricate ejective \textit{p'} [p'], \textit{t'} [t'], \textit{ch'}  [ch'] and \textit{k'} [k']. Like Spanish and Quechua it features the phonemes /\textbottomtiebar{t\textesh}/ \textit{ch}, /\textltailn/ \textit{\~n}, and /\textturny/ \textit{ll} \cite{aymara-escritura}.

\subsection{Shipibo-Konibo}

The Shipibo-Konibo people are one of the most influential communities in the Peruvian Amazon. They call themselves "Jonikon", which means "real people"; they also adopted the exonym "shipibo". Their own language or 'joikon', 'true language' is now known as Shipibo-Konibo. This language belongs to the Panoan linguistic family, which is an important subject of study for many linguistic researchers in Peru \cite{adelaar2014endangered,zariquiey2006reinterpretacion}. Shipibo-Konibo is an agglomerative language, with a high use of common suffixes (130) plus some prefixes (13) for its word-formation process. Furthermore, the basic sentence order is SOV (subject-object-verb) as opposed to Spanish (SVO) \cite{valenzuela2003transitivity}. This language is spoken by around 22 thousand people in 150 communities and is taught in almost 300 public schools \cite{minedu2018}. The majority of the population is bilingual, meaning they speak Shipibo-Konibo and Spanish. Although Shipibo-Konibo is still transmitted to children, there is a growing number of people who speak Spanish as a dominant language and achieve only partial or passive mastery of their native language. Furthermore, the degree of impact of Spanish speech and structure on Shipibo-Konibo is considerable. For these reasons, the language is considered to be in a vulnerable situation.

The phonological repertoire of Shipibo consists of 16 consonants and 4 vowels.

The vowels in Shipibo are characterized by the presence of two heights (high and low), among which it is important to point out the high central vowel, not rounded \textit{\textbari}. In the consonant phonemes we find four labial [p], [b], and [m], nine coronal [t], [s], [ts], [n], [\textesh], [\textbottomtiebar{t\textesh}], [y], [ş] and [r], two dorsal [k] and [w] and one global [h] \cite{martinez2009velarizacion}.

\section{Corpus Creation}

\subsection{Methodology}

Like Common Voice, we used the crowdsourcing method, which is based on the massive help of volunteers for audio recordings. This methodology allowed us to collect as many audios as possible in a short time and with a small budget. We used two corpus collection applications (Huqariq, Tarpuriq) designed exclusively to record and validate respectively. Unlike the Common Voice platform, the volunteers do not have to read a sentence but listen to it. This last functionality is important for native languages of Peru, due to a large part of the native speaker population are illiterate.

\subsection{Text Corpus}

This section describes the steps followed to collect the text to be used in the corpus. 

The official dictionaries of each language described in this research were used. These dictionaries are publicly available on the Internet. We used the official dictionaries issued by the Peruvian Ministry of Education, because the texts in the dictionaries are correctly written according to the official standard of each language. In Table \ref{1} we can observe the dictionaries used for the creation of our corpus.

\begin{table*}[ht]
\begin{center}
\begin{tabular}{l|c|c}
\hline
Language    & Dictionary & Year \\ \hline
Southern Quechua & Yachakuqkunapa Simi Qullqa (Chanka and Collao)     & 2005 \\
Central Quechua  & Chawpi Qichwapa Chimi Qullqan                                & 2017 \\
Aymara                            & Yatiqirinaka Aru Pirwa                                       & 2005 \\ 
Shipibo-Konibo                    & Diccionario Shipibo-Español                                   & 1993 \\ \hline
\end{tabular}
\end{center}
\caption{Dictionaries used for the construction of the corpus.}
\label{1}
\end{table*}

In order to organize the data of the collected dictionaries, a table was created manually with the following columns: Language, family, variety, region, author, dictionary name, year, lexical entry, grammatical category, gloss, definition in Spanish, definition in source language, synonym in Spanish, synonyms in source language, notes (clarifications), example in Spanish and example in source language. This table contains all the data from the dictionaries collected. This table was very helpful for the linguists who supported us in the project, since they could make filters to be able to review the data in a simpler and faster way. Finally, the entries that did not have an example in the source language were eliminated, since these examples are used as transcriptions in the corpus.

\subsection{Preprocessing and Normalization}

After obtaining all the data from the dictionaries of the different languages in a table, we eliminated all the sentences in the "example in source language" column that had more than 10 words. This was done so that the volunteers would not have problems remembering the sentence to repeat when recording their voices.

Subsequently, 4 native speaker linguists corrected, normalized and standardized the sentences in the "example in source language" column according to the grammar issued by the Ministry of Education and Ministry of Culture for each language. On the other hand, for the Southern Quechua sentences, a morphological analyzer \cite{rios2015basic} was used, which automatically standardizes according to the rules of the Ministries of Education and Culture. Table \ref{tab:oraciones} shows the number of sentences we selected from each language.  

\begin{table}[ht]
\begin{center}
\begin{tabular}{|l|c|}
\cline{1-2}
Language & Number of Sentences \\ \hline
\begin{tabular}[c]{@{}l@{}}Southern Quechua\\  (Chanka and Collao) \end{tabular}   & 8000 \\ \hline
Central Quechua & 1171\\ \hline
Aymara &  1900 \\ \hline
Shipibo-Konibo & 500 \\ \hline

\end{tabular}
\end{center}
\caption{Number of sentences for each language used in the construction of the corpus.}
\label{tab:oraciones}
\end{table}

\subsection{Recording of prompts}

Linguists who are native speakers of each language recorded their voices reading each of the selected sentences. The recordings were made using the Tarpuriq application for Android, which has an audio recording module very similar to Huqariq application for Android \cite{camacho2020language}. The recordings were made in a controlled environment, mainly free of noise and interference of any kind. All recordings made by the linguists were stored in a folder called "prompts" and folders named after their respective languages. All the recordings (prompts) made by the linguists are then entered into the Huqariq application so that they can be listened to by the volunteers to record their voices. 

Finally, the prompts were saved as 16-bit, single-channel WAV audio files with a sampling frequency of 16 kHz.

\subsection{Recording and validation of audios}

For the collection of recordings (audio files) from native speakers (users), Huqariq was used. This application allowed native speakers to record their voices repeating the sentences they hear in the prompts mentioned above. The app assigns 200 sentences per user, this feature of Huqariq was developed in this research in order for users to have a goal and to be able to be rewarded when they achieve it.

The recordings of the volunteers have the same technical information as the prompts. The recordings made by users were validated using 2 methods. The first method used an automated quality validation module that checks the noise, silence and duration of the recordings, this method was incorporated into the Huqariq application. The second method was performed by Tarpuriq, which allowed native linguists of the respective languages to validate the quality of the recordings through a voting system, this method is similar to the one used by Common Voice. Each recording must be voted 3 times, if a recording receives two positive votes, it will be marked as valid, on the contrary, if it receives two negative votes, it will be marked as invalid. Recordings marked as valid will be added to the final training, development and test corpus. These 2 methods allow to have a good quality corpus.

The validated recordings were stored in a folder where they were subsequently divided into three data sets (train, dev, test) according to statistical power analyses. Given the total number of validated recordings in a language, the number of recordings in the test set is equal to the number needed to achieve a 99\% confidence level with a margin of error of 1\% relative to the number of recordings in the training set. The same is true for the development set. 

Table \ref{3} shows the number of hours recorded and validated for each language. As can be seen, Southern Quechua has the highest number of hours collected. This is due to the fact that Southern Quechua has the largest number of native speakers compared to the other languages in this study. In addition, it has a greater participation in revitalization tasks due to the majority of research carried out for this language. Central Quechua, Aymara and Shipibo-Konibo, on the other hand, unfortunately have little or no participation in revitalization or cultural promotion.

This corpus is a July 2021 version of the corpus, which is the most updated, since due to the pandemic we have not been able to continue working on the validation of the corpus. The corpus currently has a private license, since part of the work was done with funds from private entities. For this reason, those interested can write to us if they wish to make use of it. On the other hand, the corpus statistics are visible on the Siminchikkunarayku\footnote{\url{www.siminchikkunarayku.pe}} page where the following information can be seen: language, phrase, votes, gender, accent. This information is relevant for different types of research and for that reason we consider it useful to place it, as well as in Table \ref{4} the number of hours divided by each language.

\begin{table}[ht]
\begin{center}

\begin{tabular}{|l|c|cc|}
\hline
\multirow{2}{*}{Language}                                   & \multirow{2}{*}{volunteers} & \multicolumn{2}{c|}{Hours}             \\ \cline{3-4} 
                                                            &                             & \multicolumn{1}{c|}{Total} & validated \\ \hline
\begin{tabular}[c]{@{}l@{}}Southern \\ Quechua\end{tabular} & 480                         & \multicolumn{1}{c|}{340}   & 180       \\ \hline
\begin{tabular}[c]{@{}l@{}}Central\\ Quechua\end{tabular}   & 20                          & \multicolumn{1}{c|}{20}    & 20        \\ \hline
Aymara                                                      & 8                           & \multicolumn{1}{c|}{15}    & 14        \\ \hline
Shipibo                                                     & 2                           & \multicolumn{1}{c|}{7}     & 6         \\ \hline
\end{tabular}
\end{center}
\caption{Huqariq current data statistics. This data is from the July 2021 version. }
\label{3}
\end{table}

\begin{table}[ht]
\begin{center}

\begin{tabular}{|l|ccc|}
\hline
\multirow{2}{*}{Language} & \multicolumn{3}{c|}{Hours}                                                        \\ \cline{2-4} 
                          & \multicolumn{1}{c|}{Train} & \multicolumn{1}{c|}{Dev} & \multicolumn{1}{l|}{Test} \\ \hline
Southern Quechua          & \multicolumn{1}{c|}{144}   & \multicolumn{1}{c|}{18}  & 18                        \\ \hline
Central Quechua           & \multicolumn{1}{c|}{16}    & \multicolumn{1}{c|}{2}   & 2                         \\ \hline
Aymara                    & \multicolumn{1}{c|}{12}    & \multicolumn{1}{c|}{1}   & 1                         \\ \hline
Shipibo                   & \multicolumn{1}{c|}{4}     & \multicolumn{1}{c|}{1}   & 1                         \\ \hline
\end{tabular}
\end{center}
\caption{Statistics of the number of hours divided according to train, dev and test and by each language}
\label{4}
\end{table}

\section{Automatic Speech Recognition Experiments}

The following experiment demonstrates the potential of the Huqariq corpus for multilingual speech research for low-resource languages.

For this experiment we used the corpus described in Table \ref{4}. We used the pre-trained model Wav2Vec2 \cite{baevski2020wav2vec} which was trained with 600 hours of Spanish\footnote{\url{https://huggingface.co/facebook/wav2vec2-large-xlsr-53-spanish}}. It is important to mention that we use a pre-trained model of Spanish because the languages in our corpus contain many borrowings from Spanish and this can improve the performance of the model. Moreover, we used the training setup from the public repository of which we obtained the pre-trained model. We trained our models on a GPU with 8 GB of memory for about 24 hours. In addition, we used the Adam optimizer, a learning rate of $4x10^{-5}$ and chose the Wav2letter++ decoder to obtain LM-biased results \cite{pratap2019wav2letter++}. For the other four languages the modeling units are determined by the BPE algorithm, as in \cite{zhou2018multilingual}. For the experiments, we add an additional projection layer and fit the ASR model with CTC loss as \cite{yi2020applying}. During decoding, 5-gram models are used, each of which is trained with the corresponding training transcripts. 

Table \ref{5} shows the results of the Wav2Vec2 model for each trained language and the results of previous work. The character error rate (CER) of the resulting model in the test set, defined as the Levenshtein distance \cite{fiscus2006multiple} of characters between the true transcription and the decoding result was used to measure the performance of the models for each language. It can be seen from Table \ref{5} that the Wav2Vec2 model does not outperform the previous work for Southern Quechua, this lead to the assumption that the amount of corpus used to train the Wav2Vec2 model is not large enough for the decoder to generalize well. On the other hand, the results for the other languages cannot be compared since in their case for the first time the decoder has been able to generalize well.

\begin{table*}[ht]
\begin{center}
\begin{tabular}{l|cccc}
\hline
\multicolumn{1}{l|}{Model}                                            & \begin{tabular}[c]{@{}l@{}}Southern \\ Quechua\end{tabular}          & \begin{tabular}[c]{@{}c@{}}Central \\ Quechua\end{tabular}      
& Aymara                                                   
& Shipibo                                                  \\ \hline
\begin{tabular}[c]{@{}l@{}}wav2letter++\\ + (DA)\end{tabular}                            & \begin{tabular}[c]{@{}c@{}}31.48\\ \textbf{22.75}\end{tabular}    & -                                                       & -                                                        & -                                                        \\ \hline
\begin{tabular}[c]{@{}l@{}}Wav2Vec2\\  + CTC (subword)\\ + LM (decode)\end{tabular} & \begin{tabular}[c]{@{}c@{}}28.73\\ \\ 23.19\end{tabular} & \begin{tabular}[c]{@{}c@{}}41.15\\ \\ \textbf{36.37}\end{tabular} & \begin{tabular}[c]{@{}c@{}}59.81\\ \\ \textbf{52.6}\end{tabular} & \begin{tabular}[c]{@{}c@{}}72.15\\ \\ \textbf{67.47}\end{tabular} \\ \hline
\end{tabular}
\end{center}
\caption{Performance results of the ASR models performed for each language using the CER metric.}
\label{5}
\end{table*}

\section{Concluding remarks}

We have presented Huqariq: a multilingual speech corpus of Peruvian native languages for the development of speech recognition tools. By using the crowdsourcing methodology and 2 mobile applications we have collected the largest speech corpus of native Peruvian languages. In addition, we have made some modifications to the collection applications so that they are better adapted to the problems of poorly resourced and endangered languages. We are going to release a Creative Commons CC0 licensed version so that the corpus can be in the public domain. On the other hand, we have conducted some experiments on automatic multilingual speech recognition with the Huqariq corpus using the Wav2Vec2 model.  This is the first time that speech recognition experiments have been performed for Central Quechua, Aymara and Shipibo-Konibo. Finally, we are working toward the goal that by the end of 2022 Huqariq will be able to work with 20 native languages of Peru and that many more native speakers of these languages will become volunteers.

\section{Acknowledgments}

We thank all the volunteers who are native speakers of these beautiful languages native to Peru for their time, and especially the volunteer researchers who find new texts to translate and add to the application. Special thanks to Roger Gonzalo, Virginia Mamani, and Abel Anccalle for their work on Huqariq, and all the members of the Siminchikkunarayku team.

This work was supported by the Pontifical Catholic University of Peru in 2020-2021.

\section{Bibliographical References}\label{reference}

\bibliographystyle{lrec2022-bib}
\bibliography{lrec2022-example}

\begin{thebibliography}{}

\bibitem[\protect\citename{Adelaar}2014]{adelaar2014endangered}
Adelaar, W. F.~H.
\newblock (2014).
\newblock Endangered languages with millions of speakers: Focus on quechua in
  peru.
\newblock {\em JournaLIPP 3, 2014, 1-12}.

\bibitem[\protect\citename{Ardila \bgroup et al.\egroup
  }2019]{ardila2019common}
Ardila, R., Branson, M., Davis, K., Henretty, M., Kohler, M., Meyer, J.,
  Morais, R., Saunders, L., Tyers, F.~M., and Weber, G.
\newblock (2019).
\newblock Common voice: A massively-multilingual speech corpus.
\newblock {\em arXiv preprint arXiv:1912.06670}.

\bibitem[\protect\citename{Baevski \bgroup et al.\egroup
  }2020]{baevski2020wav2vec}
Baevski, A., Zhou, H., Mohamed, A., and Auli, M.
\newblock (2020).
\newblock wav2vec 2.0: A framework for self-supervised learning of speech
  representations.
\newblock {\em arXiv preprint arXiv:2006.11477}.

\bibitem[\protect\citename{Camacho and Zevallos}2020]{camacho2020language}
Camacho, L. and Zevallos, R.
\newblock (2020).
\newblock Language technology into high schools for revitalization of
  endangered languages.
\newblock In {\em 2020 IEEE XXVII International Conference on Electronics,
  Electrical Engineering and Computing (INTERCON)}, pages 1--4. IEEE.

\bibitem[\protect\citename{Cardenas \bgroup et al.\egroup
  }2018]{cardenas2018siminchik}
Cardenas, R., Zevallos, R., Baquerizo, R., and Camacho, L.
\newblock (2018).
\newblock Siminchik: A speech corpus for preservation of southern quechua.
\newblock {\em ISI-NLP 2}, page~21.

\bibitem[\protect\citename{Duan \bgroup et al.\egroup }2019]{duan2019resource}
Duan, M., Fasola, C., Rallabandi, S.~K., Vega, R.~M., Anastasopoulos, A.,
  Levin, L., and Black, A.~W.
\newblock (2019).
\newblock A resource for computational experiments on mapudungun.
\newblock {\em arXiv preprint arXiv:1912.01772}.

\bibitem[\protect\citename{Fiscus \bgroup et al.\egroup
  }2006]{fiscus2006multiple}
Fiscus, J.~G., Ajot, J., Radde, N., Laprun, C., et~al.
\newblock (2006).
\newblock Multiple dimension levenshtein edit distance calculations for
  evaluating automatic speech recognition systems during simultaneous speech.
\newblock In {\em LREC}, pages 803--808. Citeseer.

\bibitem[\protect\citename{Heggarty \bgroup et al.\egroup }2005]{heggarty}
Heggarty, P., Valko, M.~L., Huarcaya, S.~M., Jerez, O., Pilares, G., Paz,
  E.~P., Noli, E., and Usandizaga, H.
\newblock (2005).
\newblock Enigmas en el origen de las lenguas andinas: aplicando nuevas
  t{\'e}cnicas a las inc{\'o}gnitas por resolver.
\newblock {\em Revista Andina}, 40:9--57.

\bibitem[\protect\citename{INEI}2017]{INEI}
INEI.
\newblock (2017).
\newblock Instituto nacional de estadística e informática.
\newblock
  \url{https://www.inei.gob.pe/media/MenuRecursivo/publicaciones_digitales/Est/Lib1544/}.
\newblock Accessed: 2022-07-12.

\bibitem[\protect\citename{Landerman}1992]{landerman}
Landerman, P.~N.
\newblock (1992).
\newblock Quechua dialects and their classification.
\newblock {\em PhD Thesis}.

\bibitem[\protect\citename{Martinez}2009]{martinez2009velarizacion}
Martinez, R.~R.
\newblock (2009).
\newblock La velarizaci{\'o}n en shipibo.
\newblock {\em Escritura y pensamiento}, 12(24):91--134.

\bibitem[\protect\citename{MINEDU}2018]{minedu2018}
MINEDU.
\newblock (2018).
\newblock Documento nacional de lenguas originarias del per{\'u}.
\newblock
  \url{https://centroderecursos.cultura.pe/sites/default/files/rb/pdf/Documento\%20Nacional\%20de\%20Lenguas\%20Originarias\%20del\%20Peru.pdf}.
\newblock Accessed: 2022-07-12.

\bibitem[\protect\citename{MINEDU}2021a]{aymara-escritura}
MINEDU.
\newblock (2021a).
\newblock Aymara arutha chiqapa qillqa{\~n}ataki panka= manual de escritura
  aimara.
\newblock {\em Ministerio de Educaci{\'o}n}.

\bibitem[\protect\citename{MINEDU}2021b]{QC-escritura}
MINEDU.
\newblock (2021b).
\newblock Chawpi qichwata alli qillqanapaq maytu 2. manual de escritura en
  lengua originaria quechua central.
\newblock \url{https://de.es1lib.org/book/17038295/d397c3}.
\newblock Accessed: 2022-07-12.

\bibitem[\protect\citename{Pratap \bgroup et al.\egroup
  }2019]{pratap2019wav2letter++}
Pratap, V., Hannun, A., Xu, Q., Cai, J., Kahn, J., Synnaeve, G., Liptchinsky,
  V., and Collobert, R.
\newblock (2019).
\newblock Wav2letter++: A fast open-source speech recognition system.
\newblock In {\em ICASSP 2019-2019 IEEE International Conference on Acoustics,
  Speech and Signal Processing (ICASSP)}, pages 6460--6464. IEEE.

\bibitem[\protect\citename{Rios and Mamani}2014]{rios2014morphological}
Rios, A. and Mamani, R.~C.
\newblock (2014).
\newblock Morphological disambiguation and text normalization for southern
  quechua varieties.
\newblock {\em COLING 2014}, page~39.

\bibitem[\protect\citename{Rios}2015]{rios2015basic}
Rios, A.
\newblock (2015).
\newblock {\em A basic language technology toolkit for Quechua}.
\newblock {Ph.D.} thesis, University of Zurich.

\bibitem[\protect\citename{Rogers and Campbell}2015]{rogers2015endangered}
Rogers, C. and Campbell, L.
\newblock (2015).
\newblock Endangered languages.
\newblock
  \url{https://oxfordre.com/linguistics/view/10.1093/acrefore/9780199384655.001.0001/acrefore-9780199384655-e-21}.
\newblock Accessed: 2022-07-12.

\bibitem[\protect\citename{Torero}1964]{torero}
Torero, A.
\newblock (1964).
\newblock Los dialectos quechua.
\newblock {\em Separata de Anales Científicos de la Universidad Agraria Vol.
  II}.

\bibitem[\protect\citename{Valenzuela}2003]{valenzuela2003transitivity}
Valenzuela, P.~M.
\newblock (2003).
\newblock {\em Transitivity in shipibo-konibo grammar}.
\newblock University of Oregon.

\bibitem[\protect\citename{VoxForge}2019]{VoxForge}
VoxForge.
\newblock (2019).
\newblock Voxforge.
\newblock \url{http://www.voxforge.org/}.

\bibitem[\protect\citename{Yi \bgroup et al.\egroup }2020]{yi2020applying}
Yi, C., Wang, J., Cheng, N., Zhou, S., and Xu, B.
\newblock (2020).
\newblock Applying wav2vec2. 0 to speech recognition in various low-resource
  languages.
\newblock {\em arXiv preprint arXiv:2012.12121}.

\bibitem[\protect\citename{Zariquiey and
  others}2006]{zariquiey2006reinterpretacion}
Zariquiey, R. et~al.
\newblock (2006).
\newblock Reinterpretaci{\'o}n fonol{\'o}gica de los pr{\'e}stamos l{\'e}xicos
  de base hispana en la lengua.
\newblock {\em Bolet{\'\i}n de la Academia Peruana de la Lengua}.

\bibitem[\protect\citename{Zhou \bgroup et al.\egroup
  }2018]{zhou2018multilingual}
Zhou, S., Xu, S., and Xu, B.
\newblock (2018).
\newblock Multilingual end-to-end speech recognition with a single transformer
  on low-resource languages.
\newblock {\em arXiv preprint arXiv:1806.05059}.

\end{thebibliography}

\end{document}